\documentclass[letterpaper]{article} % DO NOT CHANGE THIS
\usepackage{aaai25}  % DO NOT CHANGE THIS
\usepackage{times}  % DO NOT CHANGE THIS
\usepackage{helvet}  % DO NOT CHANGE THIS
\usepackage{courier}  % DO NOT CHANGE THIS
\usepackage[hyphens]{url}  % DO NOT CHANGE THIS
\usepackage{graphicx} % DO NOT CHANGE THIS
\usepackage{natbib}  % DO NOT CHANGE THIS AND DO NOT ADD ANY OPTIONS TO IT
\usepackage{caption} % DO NOT CHANGE THIS AND DO NOT ADD ANY OPTIONS TO IT
\usepackage{algorithm}
\usepackage{algorithmic}

\usepackage{newfloat}
\usepackage{listings}
\DeclareCaptionStyle{ruled}{labelfont=normalfont,labelsep=colon,strut=off} % DO NOT CHANGE THIS
\floatstyle{ruled}
\newfloat{listing}{tb}{lst}{}
\floatname{listing}{Listing}
\usepackage{amsmath}
\usepackage{amssymb}
\usepackage{booktabs}
\usepackage{adjustbox}
\usepackage{multirow}
\usepackage{pifont}
\usepackage{xcolor}
\usepackage{bm}
\usepackage{subcaption}
\newcommand{\ie}{\emph{i.e.}}

\title{Super-class guided Transformer for Zero-Shot Attribute Classification}

\author{
    Sehyung Kim\equalcontrib, 
    Chanhyeong Yang\equalcontrib, 
    Jihwan Park, 
    Taehoon Song, 
    Hyunwoo J. Kim\thanks{Corresponding author.}
}
\affiliations{
    Department of Computer Science and Engineering, Korea University \\
    \texttt{\{shkim129, 0814gerrardso, jseven7071, taehoons, hyunwoojkim\}@korea.ac.kr}
}

\usepackage{bibentry}
\begin{document}

\maketitle

\begin{abstract}
Attribute classification is crucial for identifying specific characteristics within image regions.
Vision-Language Models (VLMs) have been effective in zero-shot tasks by leveraging their general knowledge from large-scale datasets.
Recent studies demonstrate that transformer-based models with class-wise queries can effectively address zero-shot multi-label classification.
However, poor utilization of the relationship between seen and unseen attributes makes the model lack generalizability.
Additionally, attribute classification generally involves many attributes, making maintaining the model’s scalability difficult.
To address these issues, we propose \textbf{Su}per-class \textbf{g}uided tr\textbf{a}ns\textbf{Former} (\textbf{SugaFormer}), a novel framework that leverages super-classes to enhance scalability and generalizability for zero-shot attribute classification.
SugaFormer employs Super-class Query Initialization (SQI) to reduce the number of queries, utilizing common semantic information from super-classes, and incorporates Multi-context Decoding (MD) to handle diverse visual cues.
To strengthen generalizability, we introduce two knowledge transfer strategies that utilize VLMs. 
During training, Super-class guided Consistency Regularization (SCR) aligns model’s features with VLMs using super-class guided prompts, and during inference, Zero-shot Retrieval-based Score Enhancement (ZRSE) refines predictions for unseen attributes.
Extensive experiments demonstrate that SugaFormer achieves state-of-the-art performance across three widely-used attribute classification benchmarks under zero-shot, and cross-dataset transfer settings. Our code is available at \texttt{https://github.com/mlvlab/SugaFormer}.
\end{abstract}

\begin{figure}[ht]
    \centering
    \includegraphics[width=0.98\linewidth]{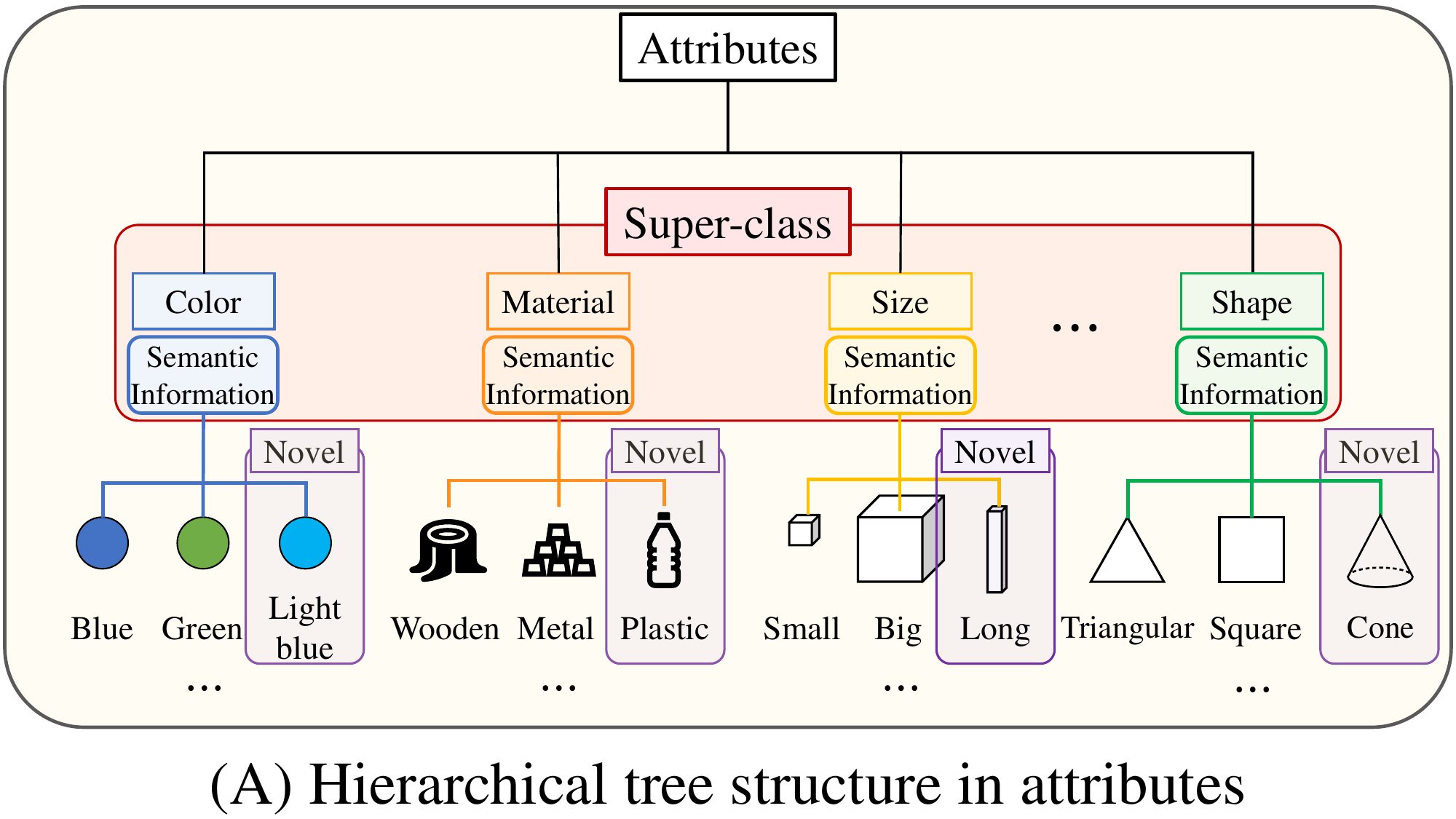}
    \includegraphics[width=0.98\linewidth]{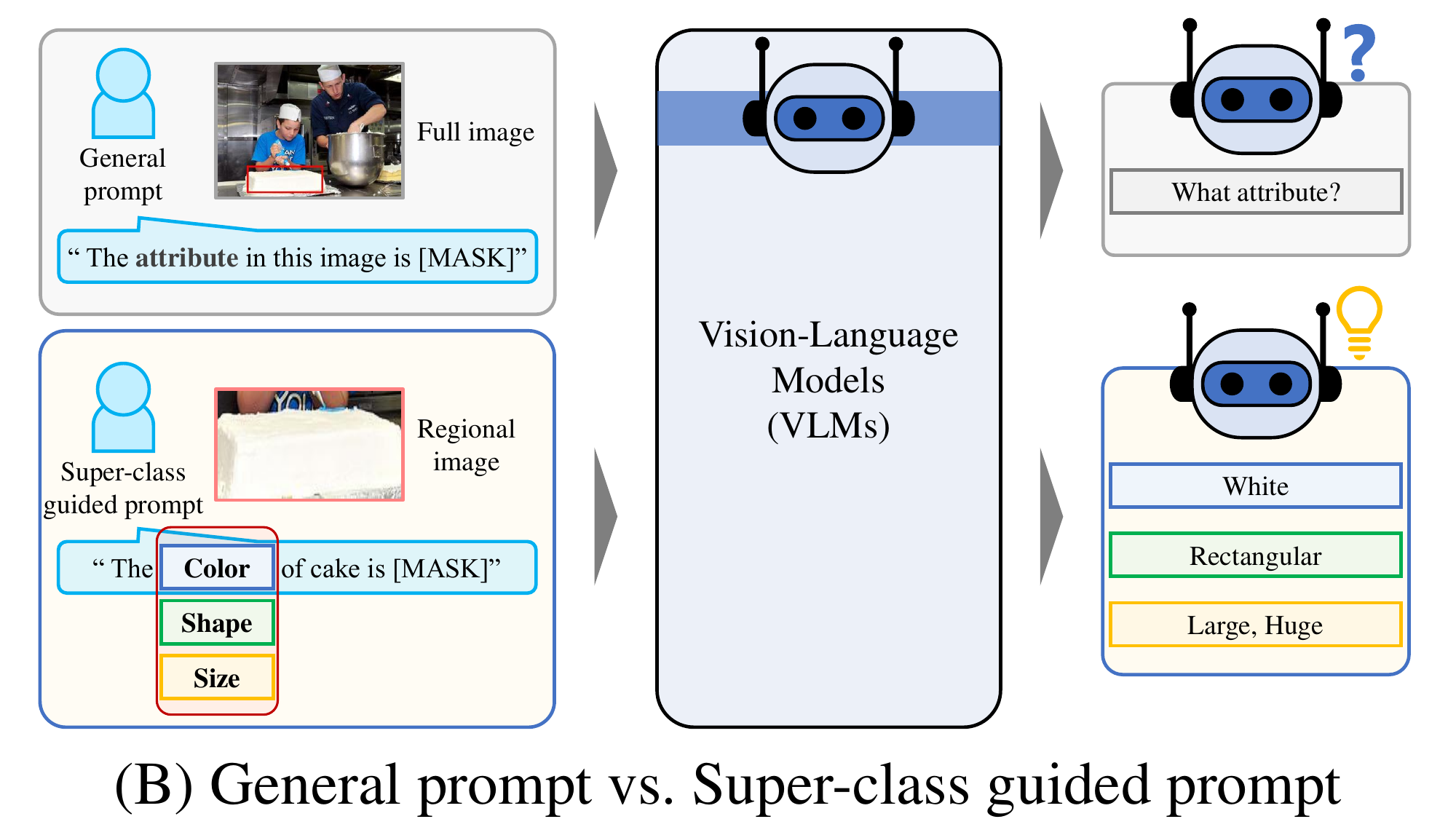}
    \caption{
    \textbf{Hierarchical structure of attributes and effectiveness of super-class guided prompt.} (A) Attributes belonging to the same super-class share common semantic information.  (B) By leveraging super-class guided prompts, VLMs make more accurate predictions by distinguishing attributes within each super-class.}
    \label{fig:common_property_in_sc}
\end{figure}

\section{Introduction}
Understanding an image at the region level is crucial for advancing computer vision. Building on the progress of image-level multi-label classification, region-level attribute classification takes this further by recognizing specific attributes within image regions. For example, it can describe a `bus' as red, large, and parking, or the `road' as wet and crowded. This task is vital for applications such as self-driving cars and image recommender systems, where the precise recognition of regional attributes directly impacts performance.

In recent years, the extension of visual recognition to zero-shot learning~\cite{xian2018zero, chen2023duet}, where models are required to predict novel classes not encountered during training, has achieved notable success~\cite{zhou2023zegclip, guo2024unseen}.
A promising strategy is to leverage pre-trained Vision-Language Models (VLMs)~\cite{radford2021learning, li2023blip}. 
This can be achieved through techniques such as fine-tuning VLMs on specific datasets, applying knowledge distillation~\cite{yang2023multilabel, liao2022gen}, or directly integrating these models into the architecture~\cite{gu2021open, ning2023hoiclip}. 
Given these insights, we leverage VLMs for zero-shot attribute classification, focusing on developing a effective and scalable approach to handle unseen classes.

One practical approach for zero-shot multi-label classification is a transformer-based classifier equipped with class-wise queries~\cite{ridnik2023ml}. 
This technique demonstrates strong performance in fine-grained classification and offers the flexibility to predict unseen labels. 
However, this line of work has two notable disadvantages. 
First, it lacks scalability for handling many classes, such as in attribute classification, since class-wise queries require substantial memory and computational cost as the number of classes increases. 
Second, the generalization of unseen classes is suboptimal, mainly due to the poor utilization of the relationship between seen and unseen classes. 

To this end, we propose \textbf{Su}per-class \textbf{g}uided tr\textbf{a}ns\textbf{Former} (\textbf{SugaFormer}), a framework that leverages super-classes to effectively utilize VLMs to address the challenges of zero-shot attribute classification. SugaFormer improves scalability and generalizability through Super-class Query Initialization (SQI), as shown in Fig.~\ref{fig:common_property_in_sc}-(A), which illustrates the hierarchical relationship between super-classes and attribute classes.
By leveraging this hierarchy, SQI reduces the number of queries and utilizes common semantic information by aligning attributes with their relevant super-classes.
It also enhances performance with Multi-context Decoding (MD), which handles diverse visual cues. To strengthen generalizability, SugaFormer incorporates Super-class guided Consistency Regularization (SCR) during training, as illustrated in Fig.~\ref{fig:common_property_in_sc}-(B), aligning its features with those of VLMs through the use of super-class guided prompt. During inference, it employs Zero-shot Retrieval-based Score Enhancement (ZRSE) to refine predictions for unseen attributes by integrating similarity scores from image and text embeddings.

The effectiveness of our SugaFormer framework is validated using three attribute classification datasets: VAW~\cite{pham2021learning}, LSA~\cite{pham2022improving}, and OVAD~\cite{bravo2023open}. Our experiments and analyses demonstrate that SugaFormer achieves state-of-the-art performance in these benchmarks under zero-shot, and cross-dataset transfer settings by leveraging super-classes to accurately identify attributes and effectively VLMs.

In summary, our contributions are as follows: 
\begin{itemize}
    \item We present a Super-class guided Transformer (SugaFormer), a novel framework that improves scalability and generalizability for zero-shot attribute classification. Super-class Query Initialization (SQI) reduces the number of queries by utilizing super-classes and leveraging their common semantic information while Multi-context Decoding (MD) enhances performance by handling diverse visual cues.
    \item We propose two knowledge transfer strategies that leverage VLMs to enhance generalizability. Super-class guided Consistency Regularization (SCR) aligns its features with VLMs using super-class guided prompts during training. During inference, Zero-shot Retrieval-based Score Enhancement (ZRSE) refines predictions for unseen attributes by integrating similarity scores between image and text embeddings.
    \item Extensive experiments demonstrate that SugaFormer significantly improves scalability and generalizability, achieving state-of-the-art performance across three attribute classification benchmarks in zero-shot, and cross-dataset transfer settings.
\end{itemize}

\section{Related Work}
\noindent \textbf{Attribute classification.}
Attribute classification aims to recognize attributes such as color, shape, etc., which belong to an object. 
Unlike other vision classification tasks like multi-label classification~\cite{chua2009nus,everingham2010pascal,irvin2019chexpert,kuznetsova2020open, xu2022open, sovrasov2022combining, liu2023causality}, attribute classification needs to consider multi-context visual cues. Depending on the attribute type, the model needs to determine which visual contexts to prioritize. Consequently, despite great success in previous studies~\cite{liu2021query2label, ridnik2023ml} on multi-label classification, the necessity of a proper model for this task has arisen.
Several works~\cite{metwaly2022glidenet, chen2023ovarnet} in this field have made much progress by leveraging various context information.
For instance, GlideNet~\cite{metwaly2022glidenet} employs different feature extractors for global, local, and intrinsic features and leverages the attention mechanism using a category estimator in a two-stage manner. 
Also, OvarNet~\cite{chen2023ovarnet} exploits region proposals obtained from an RPN and utilizes extra caption data to fine-tune CLIP~\cite{radford2021learning} model to exploit more context information for attribute classification.

\noindent \textbf{VLMs and Zero-Shot Learning.}
Recently, VLMs~\cite{radford2021learning, li2023blip, alayrac2022flamingo, liu2023visual, yuan2021florence, xiao2024florence, zeng2023x}, pre-trained with large-scale image-text pairs have accomplished great advancements.
Hence, a line of work ~\cite{cao2023detecting,he2023open} incorporates VLMs by transferring their general knowledge to downstream tasks and improves their performances.
Therefore, in zero-shot attribute classification, OvarNet~\cite{chen2023ovarnet} fine-tunes CLIP~\cite{radford2021learning} for attributes with extra datasets and leverages its embeddings via prompt tuning.
However, despite the improvements, the fine-tuning method in OvarNet may adversely affect the well-learned representations of the VLMs~\cite{kumar2022fine}. 
This could prevent the model from taking full advantage of the generalizability of the VLMs acquired during pretraining. 
Instead of the fine-tuning approach, we focus on the fact that a range of work~\cite{wei2023freeze, 2024froster, lin2022frozen, kuo2022f, kim2024retrieval} has shown that utilizing embeddings from frozen VLMs can achieve substantial performance on open-vocabulary settings across a variety of downstream tasks.
Therefore, our work explores a method that leverages embeddings from frozen pre-trained VLMs.

\section{Method}
\begin{figure*}[t]
    \centering
    \includegraphics[width=0.99\textwidth]{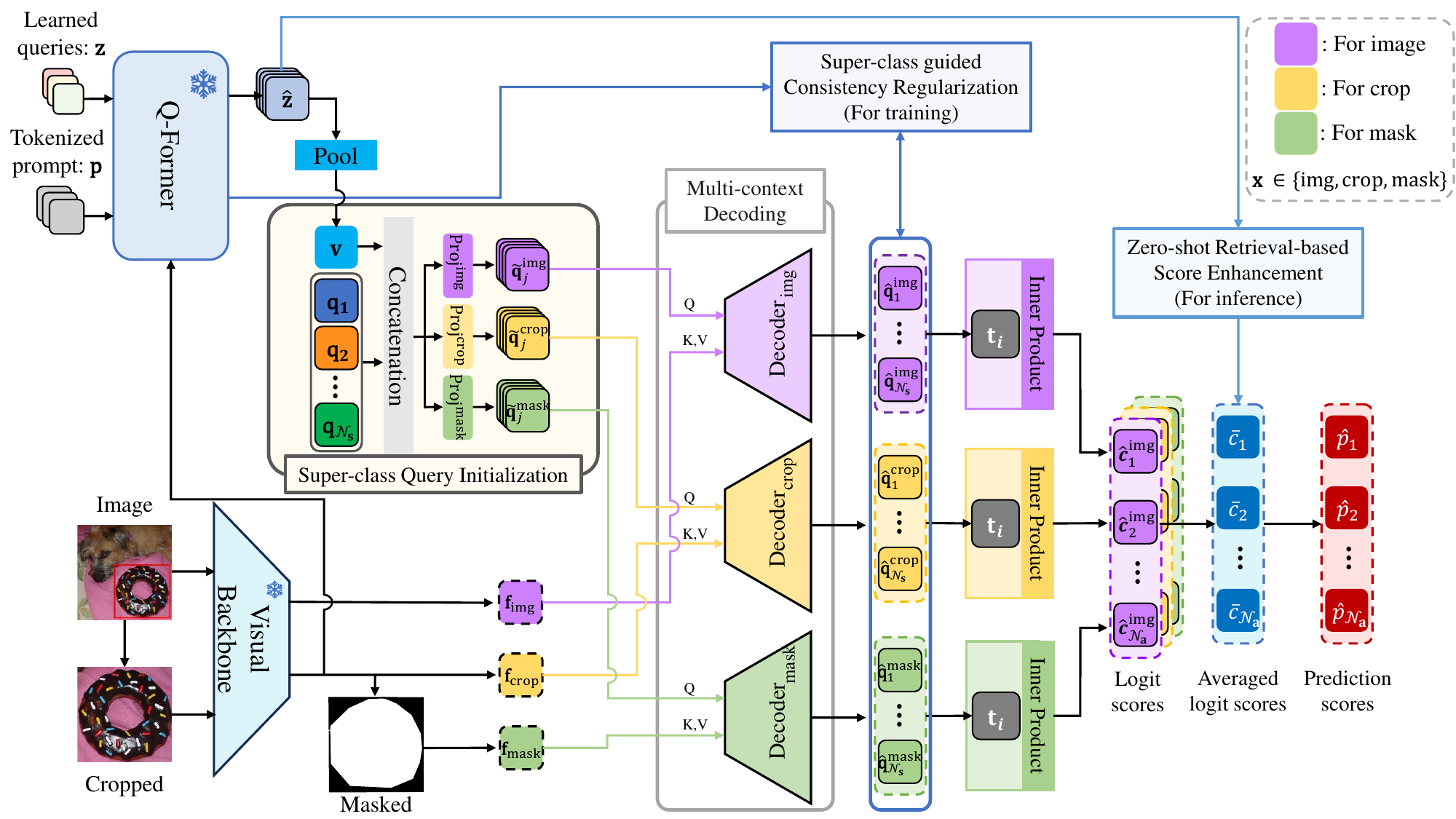}
    \caption{
    \textbf{Model architecture.} The overall pipeline of SugaFormer includes extracting multi-context visual features $\mathbf{f}_\mathbf{x}$ using
    an image, cropped image, and masked image. The super-class query $\mathbf{q}_{j}$ and pooled visual feature $\textbf{v}$ are concatenated. The concatenated query passed through different projection layers $\text{Proj}^{\text{x}}(\cdot)$, generating super-class queries $\mathbf{\tilde{q}}_j^\mathbf{x}$. Each $\text{Decoder}_{\textbf{x}}(\cdot)$ processes its respective $\mathbf{\tilde{q}}_j^\mathbf{x}$ with corresponding visual feature maps $\mathbf{f}_\mathbf{x}$, producing output queries $\mathbf{\hat{q}}^{\textbf{x}}_j$. Logit scores $\hat{c}_i^\mathbf{x}$ are computed via the inner product between $\mathbf{t}_i$ and $\mathbf{\hat{q}}^{\textbf{x}}_j$. The averaged logit score $\bar{c}_i$ is used to calculate the prediction score $\hat{p}^i$ for the $i$-th attribute. To enhance generalizability, super-class guided consistency regularization is applied during training, and zero-shot retrieval-based score enhancement is used during inference. Best viewed in color.}
    \label{fig:main_figure}
\end{figure*}

In this section, we introduce SugaFormer, a framework that leverages super-classes to enhance scalability and generalizability  for zero-shot attribute classification.
Before delving into SugaFormer, we briefly introduce the attribute classification task and base architecture in Sec.~\ref{subsec:preliminary}.
We delineate the super-class query initialization and multi-context decoding of SugaFormer in Sec.~\ref{subsec:overall architecture}.
We then present knowledge transfer strategies to leverage the VLMs for improving generalization during training and inference in Sec.~\ref{subsec:knowledge enhancement}. The overall architecture of SugaFormer is illustrated in Fig.~\ref{fig:main_figure}.

\subsection{Preliminary}
\label{subsec:preliminary}
\noindent\textbf{Problem setting.}
Attribute classification is the task of recognizing the positive attributes of an object in an image. Specifically, for each target object with bounding box $\mathbf{b} \in \mathbb{R}^{4}$, segmentation mask $\mathbf{m} \in \mathbb{R}^{\text{H}\times \text{W}}$, and object category $\mathbf{o}$ in an image $\mathbf{I} \in \mathbb{R}^{\text{H} \times \text{W} \times 3}$, an attribute classification model predicts label vectors $\mathbf{Y} \in \{1,0,-1\}^{\mathcal{N}_\mathbf{a}}$. Here, $\text{H}$ and $\text{W}$ represent the height and width of the image, $\mathcal{N}_\mathbf{a}$ is the number of attribute classes, and the values $1$, $0$, and $-1$ indicate positive, negative, and unknown attributes, respectively. In short, an attribute classifier maps a target object $(\mathbf{I}, \mathbf{b}, \mathbf{m}, \mathbf{o})$ to label vectors $\mathbf{Y}$.

In this work, we explore the zero-shot setting in attribute classification, where only a subset of attribute classes, \ie, base classes $\textbf{A}_\text{base}$, are used for training. During inference, all attribute classes $\textbf{A} = \textbf{A}{_\text{base}} \cup \textbf{A}_{\text{novel}}$ are used to test the model's generalizability to unseen attribute classes.

\noindent\textbf{Base architecture.}
We choose a transformer-based multi-label classifier ML-Decoder~\cite{ridnik2023ml} as a base model due to its performance and simple architecture.
Although this model has not been adopted for attribute classification, since this task can be seen as a multi-label classification for a target object, the model can be applied to attribution classification with minor modifications. 
For zero-shot learning, ML-Decoder uses word embeddings of attribute classes obtained from a language model~\cite{devlin-etal-2019-bert}.
Given $i$-th attribute class $\mathbf{a}_i \in \mathbf{A}$ and text encoder 
$\mathcal{T}_{\text{text}}$,  the corresponding query $\mathbf{q}_i$ is initialized by its text embedding 
$\mathbf{t}_i$ as follows:
\begin{equation}
\mathbf{q}_i = \mathbf{t}_i = \mathcal{T}_{\text{text}}(\mathbf{a}_i). 
\label{eq:txt_embedding}
\end{equation}
Then, the query $\mathbf{q}_i$ is fed into the decoder:
\begin{equation}
\mathbf{q}'_i = \textbf{CA}(\mathbf{q}_i, \mathbf{f}, \mathbf{f}) \quad \text{ and } \quad \mathbf{\hat{q}}_i = \textbf{FFN}(\mathbf{q}'_i), 
\end{equation}
where $\mathbf{f}$ is the visual feature map, $\textbf{CA}(\cdot, \cdot, \cdot)$ is a cross-attention operation and $\textbf{FFN}(\cdot)$ is a feed-forward operation. 
Finally, leveraging a output query $\mathbf{\hat{q}}$, the prediction score $\hat{p}$ is obtained as follows:
\begin{equation}
\hat{c}_i = \mathbf{t}_i \cdot \mathbf{\hat{q}}_i \quad \text{ and } \quad \hat{p}_i = \sigma(\hat{c}_i),
\end{equation}
where $\hat{c}_{i}$ is logit score for the $i$-th attribute class, and $\sigma(\cdot)$ is the sigmoid function. 
$\hat{p}_i$ is the prediction score for the $i$-th attribute class.

\subsection{Super-class guided Transformer}
\label{subsec:overall architecture}
We here delineate our \textbf{Su}per-class \textbf{g}uided tr\textbf{a}ns\textbf{Former} (\textbf{SugaFormer}) designed to enhance scalability and generalizability in zero-shot attribute classification.
SugaFormer incorporates two key components: 1) super-class query initialization, a strategy that initializes decoder queries based on super-classes to utilize their common semantic information, and 2) multi-context decoding, a scheme that decodes queries with contextual feature maps to handle diverse visual cues.

\noindent\textbf{Super-class Query Initialization.}
\label{subsec:super-class query initialization}
We propose super-class query initialization to enhance generalization capability and reduce the number of queries by leveraging the common semantic information within super-classes.
Typically, previous methods for zero-shot multi-label classification~\cite{ridnik2023ml} create one query per class.
However, the class-wise query becomes inefficient with a high number of attribute classes~\cite{pham2022improving}. 
More importantly, class-wise queries are not effective for zero-shot attribute classification training, as a single class-wise query lacks semantic information between attributes and only receives supervised signals from its corresponding class.
On the other hand, super-class queries capture the common semantic information shared by attributes within the corresponding super-class.
Hence, for more effective training, we introduce super-class queries that are shared by a group of attribute classes.
We initialize the super-class queries with the text embeddings of super-class names and visual representation of the target object for better query initialization.
We adopt Q-Former in BLIP2~\cite{li2023blip} to extract the object-conditioned visual feature $\textbf{v}\in\mathbb{R}^{d_{q}}$.
The image $\mathbf{I}$ with the target object box $\mathbf{b}$ serves as the input of the visual backbone $\mathcal{V}_{\text{img}}$, which is connected to Q-Former.
Then Q-Former extracts the output features $\mathbf{\hat{z}}\in\mathbb{R}^{\mathcal{N}_\mathbf{z}\times {d_{q}}}$ from the fixed set of learned queries $\mathbf{z}\in\mathbb{R}^{\mathcal{N}_\mathbf{z}\times {d_{q}}}$ by operating cross-attention with the feature maps from $\mathcal{V}_{\text{img}}$.
We pool the $\mathbf{\hat{z}}$ to obtain the visual features. The above process is formulated as follows:
\begin{equation}
    \begin{split}
    \mathbf{\hat{z}}&=\text{Q-Former}(\mathbf{z},\mathcal{V}_{\text{img}}(\mathbf{\Phi}_{\text{crop}}(\mathbf{I},\mathbf{b}))), \\
    \textbf{v}&=\textbf{Pool}({\mathbf{\hat{z}}}),
    \label{eq:cmq1}
    \end{split}
\end{equation}
where $\mathbf{\Phi}_\text{crop}$ represents the cropping function, and $\textbf{Pool}(\cdot)$ denotes a pooling function, such as mean or max.
Finally, we concatenate the $j$-th super-class query $\mathbf{q}_j$ with the object-conditioned visual features $\textbf{v}$ to form a combined query $\mathbf{\tilde{q}}_j$, which is then projected onto a $d_q$-dimensional feature space:
\begin{equation}
    \begin{split}
    \mathbf{\tilde{q}}_j &=\text{Proj}([\mathbf{q}_j;\textbf{v}]),
    \end{split}
    \label{eq:cmq2}
\end{equation} 
where $\text{Proj}\in\mathbb{R}^{{d\times 2d_{q}}}$ and $[\cdot ; \cdot]$ are a linear projection and concatenation, respectively.

\noindent\textbf{Multi-context Decoding.}
\label{subsec:Multi-context decoding}
We propose a decoding strategy to utilize diverse contextual information to enhance performance.
Our model first extracts three types of feature maps: cropped feature $\mathbf{f}_\text{crop}$, masked feature $\mathbf{f}_\text{mask}$, and image feature $\mathbf{f}_\text{img}$.

The cropped feature $\mathbf{f}_\text{crop}$ is extracted from the image cropped by bounding box $\mathbf{b}$ to capture the local information.
We used a large-scale pretrained visual backbone $\mathcal{V}_\text{img}$, such as CLIP~\cite{radford2021learning}, as an encoder.
To focus only on the target object (\ie, foreground), the masked feature $\mathbf{f}_\text{mask}$ is extracted after removing the background context by ground-truth segmentation mask $\mathbf{m}$.
Additionally, for incorporating the global context, the entire image is fed into the visual encoder to obtain $\mathbf{f}_\text{img}$.
The detailed formulations for extracting each feature map are provided below.
\begin{equation}
    \begin{split}
        \mathbf{f}_\text{img} \in\mathbb{R}^{hw \times d_{v}} & = \mathcal{V}_{\text{img}} (\mathbf{I}), \\
        \mathbf{f}_\text{crop}\in\mathbb{R}^{hw \times d_{v}} & = \mathcal{V}_{\text{img}} (  \mathbf{\Phi}_{\text{crop}}(\mathbf{I}, \mathbf{b}) ), \\
        \mathbf{f}_\text{mask}\in\mathbb{R}^{hw \times d_{v}} & = \mathbf{f}_\text{crop} \odot \mathbf{\Phi}_\text{resize}(\mathbf{m}),
    \label{eq:box_patch}
    \end{split}
\end{equation}
where $\mathbf{\Phi}_\text{crop}$ and $\mathbf{\Phi}_\text{resize}$ are image cropping and resizing functions, respectively, and $\odot$ denotes element-wise multiplication.
Similar to the query initialization in Eq.~\eqref{eq:txt_embedding}, 
a set of super-class queries $Q=\{\mathbf{q}_j\}_{j=1}^{\mathcal{N}_\mathbf{s}}$ are initialized by $\mathbf{q}_j = \mathbf{t}_j = \mathcal{T}_{\text{text}}(\mathbf{s}_j)$
, where $\mathbf{s}_j \in \mathbf{S}$ represents the $j$-th super-class in the super-class set $\mathbf{S} = \{\mathbf{s}_j\}_{j=1}^{\mathcal{N}_\mathbf{s}}$.
Then, multi-context decoding is performed using super-class queries $Q$ and feature maps $\mathbf{f}_\text{crop},\mathbf{f}_\text{mask},$ and $\mathbf{f}_\text{img}$.
Here, each feature map is processed independently via cross-attention with queries $Q$ for separate decoders $\text{Decoder}_\textbf{x}(\cdot, \cdot,  \cdot)$ to handle diverse visual cues, where $\textbf{x} \in$ $\{$ img, crop, mask $\}$.
Also, for each decoder, the super-class queries $Q$ are separately initialized using different linear projection $\text{Proj}^\textbf{x}$ as follows:
\begin{equation}
    \begin{split}
    \mathbf{\tilde{q}}_j^\textbf{x}&=\text{Proj}^\textbf{x}([\mathbf{q}_j;\textbf{v}]),
    \end{split}
    \label{eq:query}
\end{equation}
where $\mathbf{\tilde{q}}_j^\textbf{x}$ represents the combined query for context $\textbf{x}$.
The output features of $j$-th super-class query for each decoder can be formulated as: 
\begin{equation}
    \begin{split}
        \mathbf{\hat{q}}^\text{img}_j &= \text{Decoder}_\text{img}(\mathbf{\tilde{q}}^\text{img}_j, \mathbf{\tilde{f}}_\text{img}, \mathbf{\tilde{f}}_\text{img}), \\
        \mathbf{\hat{q}}^\text{crop}_j &= \text{Decoder}_\text{crop}(\mathbf{\tilde{q}}^\text{crop}_j, \mathbf{\tilde{f}}_\text{crop},\mathbf{\tilde{f}}_\text{crop}), \\
        \mathbf{\hat{q}}^\text{mask}_j &= \text{Decoder}_\text{mask}(\mathbf{\tilde{q}}^\text{mask}_j, \mathbf{\tilde{f}}_\text{mask}, \mathbf{\tilde{f}}_\text{mask}),
    \label{eq:mcd}
    \end{split}
\end{equation}
where $\tilde{\mathbf{f}}_\textbf{x} \in \mathbb{R}^{{hw}\times d}$ represents the projected features for each $\textbf{x}$.

\begin{figure*}[t]
    \centering
    \includegraphics[width=0.95\textwidth]{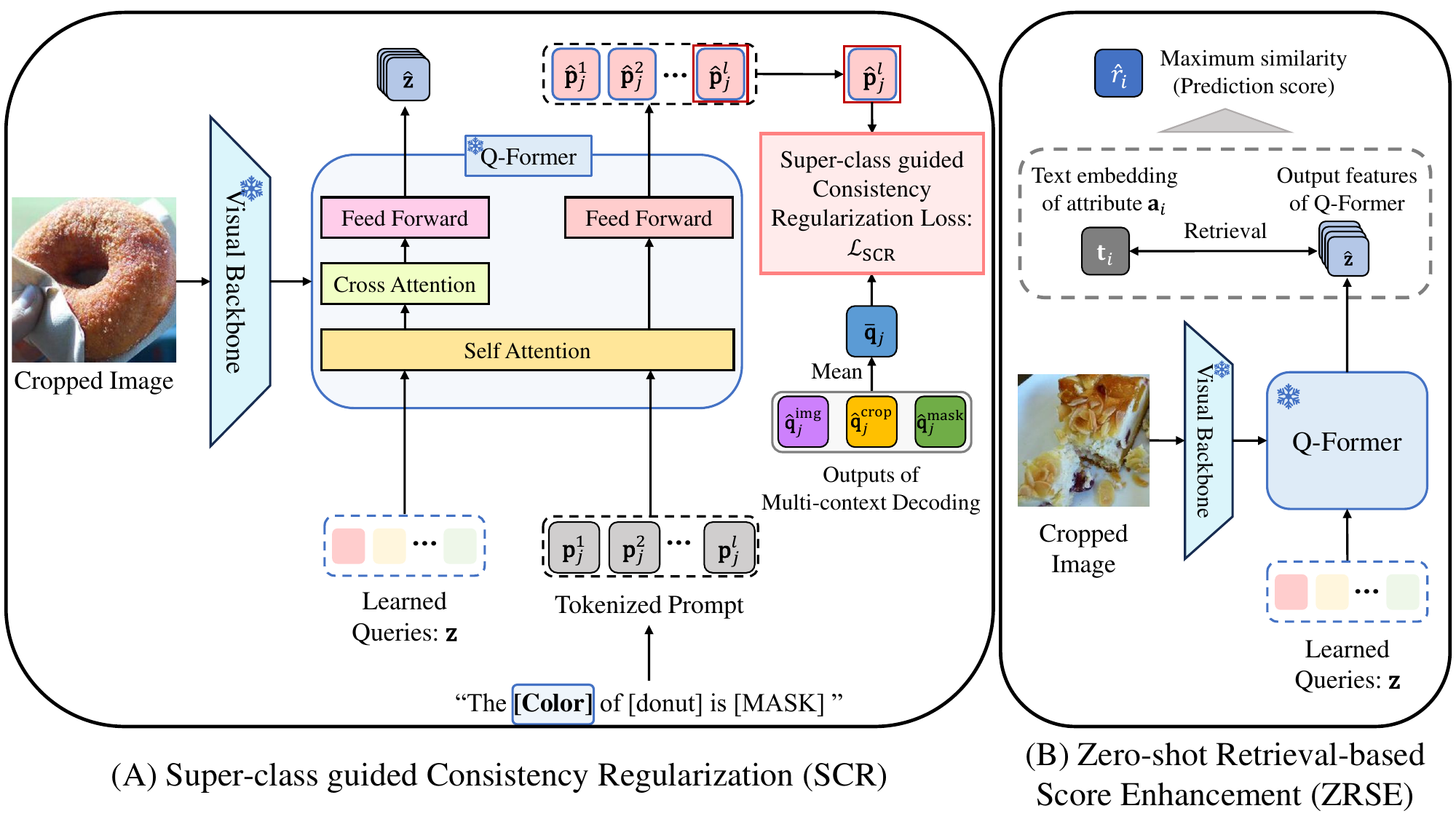}
    \caption{
    {\textbf{Illustration of knowledge transfer strategies.}}
    (A) During training, the Q-Former extracts a [MASK] token feature $\mathbf{\hat{p}}_j^l$ using a super-class guided prompt $\mathbf{p}_j$. This process leverages learned queries $\mathbf{z}$ and a tokenized prompt which is obtained by using a the prompt $\mathbf{p}_j$ that integrates the $j$-th super-class and the object class name.
    We compute $\mathcal{L}_{\text{SCR}}$ which is obtained by measuring L1 distance between the mean of output features $\mathbf{\bar{q}}_{j}$ from multi-context decoding and the [MASK] token feature $\mathbf{\hat{p}}_j^l$. 
    (B) During inference, we employ ZRSE in which  maximum similarity $\hat{r}_i$ obtained from Q-Former to compensate for novel classes.
    Note that we use the same frozen Q-Former.
    }
    \label{fig:sckd_figure}
\end{figure*}

\noindent\textbf{Attribute prediction.}
Our framework outputs the final attribute predictions by integrating all predictions from three decoders' outputs: $\mathbf{\hat{q}}^\text{img}_j, \mathbf{\hat{q}}^\text{crop}_j$, and $\mathbf{\hat{q}}^\text{mask}_j$.
For context $\mathbf{x}$ and $i$-th attribute class, the logit score $\hat{c}^{\textbf{x}}_{i}$ is computed by the the inner product between text embedding $\mathbf{t}_{i}$ and its corresponding super-class output query feature $\mathbf{\hat{q}}_j^\textbf{x}$.
Then, the final prediction score $\hat{p}_i$ for $i$-th attribute class is calculated as:
\begin{equation}
        \hat{c}_{i}^\textbf{x} = \mathbf{t}_{i}\cdot \mathbf{\hat{q}}_{j=\delta(i)}^\mathbf{x} , \text{ and }
        \hat{p}_{i} = \sigma({\bar{c}_i}),
    \label{eq:prediction}
\end{equation}
where $j=\delta(i)$ denotes the super-class index of the $i$-th attribute class and $\bar{c}_i=(\hat{c}_i^\text{img} + \hat{c}_i^\text{crop} + \hat{c}_i^\text{mask})/3$ denotes averaged logit score.

\subsection{Knowledge transfer strategies}
\label{subsec:knowledge enhancement}
In this section, we introduce knowledge transfer strategies by leveraging VLMs to improve generalizability for zero-shot attribute classification with 1) super-class guided consistency regularization and 2) zero-shot retrieval-based score enhancement.

\noindent\textbf{Super-class guided Consistency Regularization.}
We here propose Super-class guided Consistency Regularization scheme using a Image-grounded Text Generation (ITG) with a super-class guided prompt as described in Fig.~\ref{fig:sckd_figure}-(A).
Given the cropped image of the target object as input, 
our approach uses Q-Former to generate texts by employing multimodal causal masked attention~\cite{li2023blip} with the learned queries $\mathbf{z}$. 
The output feature corresponding to the prompt $\mathbf{p}$, which is tokenized into $l$ tokens, is computed as:
\begin{equation}
        \mathbf{\hat{p}} = \text{Q-Former}([\mathbf{z};\mathbf{p}],
        \mathcal{V}_{\text{img}}
        (\mathbf{\Phi}_{\text{crop}}(\mathbf{I},\mathbf{b}))),
    \label{eq:ITG}
\end{equation}
where $\mathbf{\hat{p}}\in\mathbb{R}^{l \times d_q}$ denotes the features extracted from the prompt $\mathbf{p}$ including [MASK] token.
Here, we construct a super-class guided prompt $\mathbf{p}_{j}$ for the $j$-th super-class as \texttt{`The [super-class] of the [object] is [MASK]'}.
For instance, in Fig.~\ref{fig:sckd_figure}-(A), we can use \texttt{`The [color] of the [donut] is [MASK]'} for super-class \texttt{[color]}. 
The super-class guided prompt $\mathbf{p}_j$ constructed with $j$-th super-class name encourages Q-Former to generate text tokens more relevant to attribute classification.
Instead of the generated text, we use the output feature $\mathbf{\hat{p}}^{l}_j\in\mathbb{R}^{d_q}$, where the $l$-th token corresponds to the [MASK] token from the prompt $\mathbf{p}_j$.
We compute the L1 loss between the output feature $\mathbf{\hat{p}}_j^l$ and the averaged output features $\mathbf{\bar{q}}_j=(\mathbf{\hat{q}}_j^\text{img} + \mathbf{\hat{q}}_j^\text{crop} + \mathbf{\hat{q}}_j^\text{mask})/3$ obtained from multiple decoders $\text{Decoder}_\textbf{x}$. 
The loss function for super-class guided consistency regularization $\mathcal{L}_{\text{SCR}}$ is formulated as follows:
\begin{equation}
    \mathbf{\hat{p}}_j = \text{Q-Former}([\mathbf{z};\mathbf{p}_j],         \mathcal{V}_{\text{img}}
        (\mathbf{\Phi}_{\text{crop}}(\mathbf{I},\mathbf{b}))), 
    \label{eq:KD_loss}
\end{equation}
\begin{equation}
        \mathcal{L}_{\text{SCR}} = \sum_{j=1}^{\mathcal{N}_\textbf{s}}|\mathbf{\hat{p}}_j^{l} - \mathbf{\bar{q}}_j |.
    \label{eq:KD_loss2}
\end{equation}
The total loss function of ours can be written as:
\begin{equation}
    \mathcal{L}_{\text{total}} = \sum_\textbf{x}\mathcal{L}_{\text{asym}}^\textbf{x} + \lambda\mathcal{L}_{\text{SCR}}, 
    \label{eq:loss_functions}
\end{equation}
where $\mathcal{L}_{\text{asym}}$ is an asymmetric loss function for multi-label classification~\cite{ridnik2021asymmetric}, and 
$\lambda$ denotes the weight for $\mathcal{L}_{\text{SCR}}$.

\begin{table*}[t]
  \centering
  \begin{adjustbox}{width=\textwidth}
  \begin{tabular}{ c | c | c |c c c}
    \toprule
    \textbf{Method} &
    \textbf{VLM} & 
    $\textbf{Attribute-related data for training}$ & $\textbf{AP}_{\textbf{base}}$ & $\textbf{AP}_{\textbf{novel}}$ & $\textbf{AP}_{\textbf{all}}$ \\
    \midrule
    $\text{CLIP}$~\cite{radford2021learning} &CLIP & $-$ & 50.04 & 46.54 & 49.60\\
    $\text{BLIP2}$~\cite{li2023blip} &BLIP2 & $-$ & 47.50 & 46.02 & 47.31\\
    \midrule
    $\text{CLIP-Attr}$~\cite{chen2023ovarnet} &CLIP & $\text{VAW}_{\text{base}}$ & 67.90 & 57.39 & 66.92\\
    $\text{CLIP-Attr}$~\cite{chen2023ovarnet} &CLIP & $\text{VAW}_{\text{base}}+\text{CC-3M-sub}$  & 69.79 & 59.16 & 68.87\\
    $\text{CLIP-Attr}$~\cite{chen2023ovarnet}  &CLIP & $\text{VAW}_{\text{base}}+\text{CC-3M-sub}+\text{COCO-Cap-sub}$  & 70.24 & 57.73 & 69.03\\
    $\text{OvarNet}$~\cite{chen2023ovarnet}  &CLIP & $\text{VAW}_{\text{base}}$ & 68.27 & 53.75 & 66.85\\
    $\text{OvarNet}$~\cite{chen2023ovarnet} &CLIP & $\text{VAW}_{\text{base}}+ \text{CC-3M-sub}$ & 69.30 & 55.44 & 67.96\\
    $\text{OvarNet}$~\cite{chen2023ovarnet}  &CLIP &$\text{VAW}_{\text{base}}+\text{CC-3M-sub}+\text{COCO-Cap-sub}$ & 69.80 & 56.4 & 68.52\\
    \midrule
    $\text{ML-Decoder}^\dagger$~\cite{ridnik2023ml} & BLIP2 & $\text{VAW}_{\text{base}}$ & 73.12 & 53.38 & 
    70.61\\
    $\text{SugaFormer}$ &BLIP2 &
    $\text{VAW}_{\text{base}}$ & \textbf{75.18} & \textbf{60.59} & \textbf{73.32}\\
  \bottomrule
  \end{tabular}
  \end{adjustbox}
  \caption{\textbf{Results on the VAW in zero-shot setting.} $\dagger$ denotes the results obtained from our  implementation. CC-3M-sub~\cite{changpinyo2021conceptual} and COCO-Cap-sub~\cite{lin2014microsoft} represent additional caption datasets filtered to extract attribute-related information for training.}
  \label{tab:main_zeroshot}
\end{table*}

\noindent\textbf{Zero-shot Retrieval-based Score Enhancement.}
\label{subsec:ZS}
We propose a new plug-and-play module to enhance the final predictions on unseen attribute classes by leveraging the prediction scores from Q-Former, as described in Fig.~\ref{fig:sckd_figure}-(B).
In addition to the classification score for the $i$-th attribute class $\bar{c}_i$, obtained by averaging the logit scores from the multi-context decoder, we compute the attribute classification score using Q-Former, following a process similar to Image-Text Contrastive (ITC).
In Q-Former, each output embedding $\mathbf{\hat{z}}$ generated from the learned queries $\mathbf{z}$ is individually compared to the text embedding, and the maximum similarity score is selected as the overall similarity score between the given image and text.
Similar to the above process, we first calculate the text embedding of $i$-th attribute class $\mathbf{t}_{i}$, and each $k$-th output embedding of Q-Former $\mathbf{\hat{z}}_k$, and use maximum similarity $\hat{r}_i$ as a prediction score of $i$-th attribute class from Q-Former.
This process can be formulated as:
\begin{equation}
    \begin{split}
        \hat{r}_{i} &= \underset{k=1,2,...,\mathcal{N}_{\mathbf{z}}}{\text{max}}\mathbf{t}_{i}\cdot \mathbf{\hat{z}}_k^{\top}.
    \label{eq:zs_scores}
    \end{split}
\end{equation}
Finally, we choose Top-K scores from $\mathbf{R}=\{\hat{r}_i\}_{i=1}^{\mathcal{N}_{\mathbf{a}}}$ and add it to the corresponding predictions of attribute classes from our model.
Then the Eq.~\eqref{eq:prediction} can be re-written as:
\begin{equation}
    \begin{split}
        \hat{p}_i = 
        \begin{cases}
        \sigma(\bar{c}_i + \hat{r}_i), & \text{if}\ i\in \text{Top-K}(\mathbf{R}) \\
        \sigma(\bar{c}_i), & \text{otherwise}
        \end{cases}
    \label{eq:zs_scores2}
    \end{split}
\end{equation}
where $\text{Top-K}(\mathbf{R})$ refers to the list of attribute class indexes with the Top-K similarity scores from Q-Former.

\section{Experiments}
Here, we evaluate SugaFormer using three attribute classification benchmark datasets: VAW~\cite{pham2021learning}, LSA~\cite{pham2022improving} in the zero-shot setting, and OVAD~\cite{bravo2023open} in the cross-dataset setting.
We demonstrate that SugaFormer effectively learns the semantic information shared by attributes within the same super-class, enhancing scalability and generalizability.
\subsection{Datasets}
\noindent \textbf{VAW.} VAW~\cite{pham2021learning} consists of 620 attribute classes with object instances, including segmentation masks, box coordinates, and class names. The dataset combines images from VGPhraseCut~\cite{wu2020phrasecut} and GQA~\cite{hudson2019gqa}. For zero-shot attribute classification, following prior work~\cite{chen2023ovarnet}, we use half of the `tail' attributes and 15\% of `medium' attributes as novel classes, resulting in 79 novel classes and 541 base classes. We adopt eight predefined super-classes: color, material, shape, size, texture, state, action, other.

\noindent \textbf{LSA.} LSA~\cite{pham2022improving} aggregates images and attributes from various datasets, including Visual Genome, GQA, COCO-Attributes, Flickr30K-Entities, and MS-COCO. We evaluate under zero-shot setting (common-to-rare), with 5526 base attributes and 4012 novel attributes.

\noindent \textbf{OVAD.} OVAD~\cite{bravo2023open} introduces a test-only cross-dataset benchmark featuring 117 attributes and object instances for open-vocabulary attribute detection. 
The dataset organizes attributes into three subsets: `head’ (15 frequently occurring attributes), `medium’ (53 moderately frequent attributes), and `tail’ (49 rare attributes) based on thresholds defined by attribute annotation frequency.
Experiments assume ground-truth boxes are given during testing.

\subsection{Evaluation metric}
Our evaluation metric is average precision (AP), assessed across three categories: base classes ($\text{AP}_\text{base}$), novel classes ($\text{AP}_\text{novel}$), and all attribute classes ($\text{AP}_\text{all}$).
$\text{AP}_\text{base}$ indicates the precision for ``seen'' labels present during training, reflecting its ability to recognize learned attributes. $\text{AP}_\text{novel}$ indicates the precision for ``unseen'' labels, which were not encountered during training.
$\text{AP}_\text{all}$ provides an precision of the model's performance across both seen and unseen labels.

\subsection{Implementation details}
Our model employs ViT-g/14 as its visual backbone. 
We keep all parameters within the visual backbone during training and Q-Former frozen.
We establish a baseline using ML-Decoder~\cite{ridnik2023ml} with the same visual backbone to ensure a fair comparison.
All training and evaluations are conducted on NVIDIA RTX 3090. 
Further implementation details, including hyperparameters, can be found in the supplementary materials.

\begin{table}[t]
    \centering
    \begin{adjustbox}{width=\columnwidth}
    {\fontsize{12}{14}\selectfont
    \begin{tabular}{c|c|c}
        \toprule
        \textbf{Method} & \textbf{Num. Query} & \textbf{AP}$_{\textbf{novel}}$ \\
        \midrule
        CLIP~\cite{radford2021learning}      & -     & 2.63 \\
        BLIP2~\cite{li2023blip}     & -     & 2.58 \\
        TAP~\cite{pham2022improving} & -  & 5.37 \\
        OvarNet~\cite{chen2023ovarnet}          & -  & 5.48 \\
        \midrule
        ML-Decoder~\cite{ridnik2023ml} & 9538  & out of memory \\   
        SugaFormer          & 9     & \textbf{5.80} \\
        \bottomrule
    \end{tabular}
    }
    \end{adjustbox}
    \caption{\textbf{Results on the LSA in the zero-shot setting.}}
    \label{tab:lsa_results}
\end{table}

\begin{table}[t]
  \centering
  \begin{adjustbox}{width=\columnwidth}
  {\fontsize{12}{12}\selectfont
  \begin{tabular}{ c | c |c c c}
    \toprule
    \textbf{Method}&  $\textbf{AP}_\textbf{all}$ & $\textbf{AP}_\textbf{head}$ & $\textbf{AP}_\textbf{medium}$ & $\textbf{AP}_\textbf{tail}$ \\
    \midrule
    ALBEF~\cite{li2021align}  &21.0 &44.2 &23.9 &9.4\\
    BLIP~\cite{li2022blip} &24.3 &51.0 &28.5 &9.7\\
    BLIP2~\cite{li2023blip} &25.5 &49.7 &30.4 &10.8\\
    X-VLM~\cite{zeng2021multi} &28.1 &49.7 &34.2 &\textbf{12.9}\\
    OVAD~\cite{bravo2023open} &21.4 &48.0 &26.9 &5.2\\
    CLIP-Attr~\cite{chen2023ovarnet} &26.1 &55.0 &31.9 &8.5\\
    OvarNet~\cite{chen2023ovarnet} 
    &28.6 &58.6 &35.5 &9.5\\
    \midrule
    SugaFormer 
    &\textbf{29.7} &\textbf{58.8} &\textbf{36.6}  &10.8\\
  \bottomrule
  \end{tabular}
  }
  \end{adjustbox}
  \caption{\textbf{Results on the OVAD in the cross-dataset transfer setting.}}
  \label{tab:ovad}
\end{table}

\begin{table}[t]
    \centering
    \begin{adjustbox}{width=\columnwidth}
    {\fontsize{12}{18}\selectfont
    \begin{tabular}{c c c c|c c c}
        \toprule
         \textbf{SQI} &\textbf{MD} &\textbf{SCR} &\textbf{ZRSE}           &$\textbf{AP}_\textbf{base}$ & $\textbf{AP}_\textbf{novel}$ & $\textbf{AP}_\textbf{all}$ \\
        \midrule    
         \multicolumn{4}{c}{ML-Decoder (\text{baseline})} & 73.12 &53.38 & 70.61 \\
        \midrule            
         \checkmark & & & & 72.86 & 55.90 & 70.67 \\
         \checkmark &\checkmark & & & 74.38 & 56.25 & 72.07   \\
         \checkmark &\checkmark &\checkmark & & 74.49 & 58.38 & 72.44 \\
         \checkmark &\checkmark &\checkmark  &\checkmark & \textbf{75.18} \textbf{\textcolor{red}{(+2.06)}} & \textbf{60.59} \textbf{\textcolor{red}{(+7.21)}}& \textbf{73.32} \textbf{\textcolor{red}{(+2.71)}} \\
        \bottomrule
    \end{tabular}
    }
    \end{adjustbox}
    \caption{\textbf{Ablations on key components.}}
    \label{tab:ablation_main}
\end{table}

\begin{table}[t]
    \centering
    \begin{adjustbox}{width=\columnwidth}
    {\fontsize{12}{14}\selectfont
    \begin{tabular}{c|c c c}
        \toprule
        \textbf{Prompt type} &             
        $\textbf{AP}_\textbf{base}$ & $\textbf{AP}_\textbf{novel}$ & $\textbf{AP}_\textbf{all}$ \\
        \midrule                 
         no prompt & 75.06 & 58.85 & 73.00 \\
         general prompt & 73.84 & 55.64 & 71.52 \\
         super-class guided prompt &\textbf{75.18} &\textbf{60.59} &\textbf{73.32} \\
        \bottomrule
    \end{tabular}
    }
    \end{adjustbox}
    \caption{\textbf{Analysis on regularization strategies.}}
    \label{tab:kd}
\end{table}

\subsection{State-of-the-art comparison}
\noindent \textbf{Zero-shot attribute classfication.}
To validate the effectiveness of SugaFormer, we compare it with state-of-the-art methods and our baseline.
In Tab.\ref{tab:main_zeroshot}, SugaFormer achieves an $\text{AP}_\text{novel}$ of 60.59 in the VAW zero-shot setting, demonstrating the benefits of leveraging super-classes to enhance generalizability for novel attributes.
In Tab.\ref{tab:lsa_results}, SugaFormer achieves an $\text{AP}_\text{novel}$ of 5.80 in the LSA zero-shot setting, outperforming other methods. We could not train the baseline due to out-of-memory issues caused by the many queries. These results highlight SugaFormer's superior performance in zero-shot attribute classification.

\noindent \textbf{Cross-dataset transfer.} 
In Tab.~\ref{tab:ovad}, we compare SugaFormer with previous methods on the OVAD in the cross-dataset transfer setting. Our model is trained on the VAW dataset. It is worth mentioning that training on VAW performs poorly compared to image-caption pair datasets in cross-dataset transfer, as discussed in OVAD~\cite{bravo2023open}. Nevertheless, SugaFormer achieves an $\text{AP}_\text{all}$ of 29.7, surpassing the previous best method by approximately across all categories.
\subsection{Ablation study and analysis}
\noindent \textbf{Ablation on key components.}
In Tab.~\ref{tab:ablation_main}, we conducted ablation studies on the VAW in the zero-shot setting by gradually incorporating suggested methods into our baseline model. 
When incorporating techniques, such as super-class query initialization (SQI), super-class guided consistency regularization (SCR), and zero-shot retrieval-based score enhancement (ZRSE), We observed improvements in AP$_\text{novel}$ by approximately 2.52, 2.13, and 2.21, respectively.
Moreover, multi-context decoding (MD) boosts AP$_\text{all}$, with an improvement of 1.40.
When all components were combined, the model achieved its highest performance, outperforming the baseline by 2.06 for AP$_\text{base}$, 7.21 for AP$_\text{novel}$, and 2.71 AP$_\text{all}$. These results demonstrate that leveraging super-classes enhances the generalizability of attribute classification while utilizing diverse visual cues improves overall performance.

\noindent \textbf{Analysis on regularization strategies.}
In this experiment, we explore three regularization strategies using different prompts on the VAW in the zero-shot setting.
First, the no prompt approach used only image features without any textual prompt. A caption was generated through the ITG process in Q-Former, and the resulting embedding was equally compared to all super-class queries.
Second, the general prompt without a super-class followed the format `\texttt{The attribute of the [object] is [MASK]}.' using the \texttt{[MASK]} token for consistency.
Finally, the super-class guided prompt employed the format `\texttt{The [super-class] of the [object] is [MASK].}' ensuring consistency through the use of the [MASK] token.
As shown in Tab.~\ref{tab:kd}, the best performance was achieved when prompts were tailored to each super-class, demonstrating that utilizing super-classes in prompts effectively extracts features particularly useful for attribute prediction.
% \appendix
\section{Conclusion}
We propose SugaFormer, a Super-class guided Transformer framework designed to address scalability and generalizability in zero-shot attribute classification. SugaFormer employs Super-class Query Initialization (SQI) to reduce queries by utilizing shared semantic information and Multi-context Decoding (MD) to enhance performance using diverse visual cues. Additionally, we introduce two knowledge transfer strategies leveraging VLMs: Super-class guided Consistency Regularization (SCR) aligns features during training, and Zero-shot Retrieval-based Score Enhancement (ZRSE) refines predictions during inference. Experiments on three benchmarks show that SugaFormer outperforms existing methods across zero-shot, and cross-dataset transfer settings.

\section{Acknowledgements}
This work was partly supported by the Institute of Information \& Coummunications Technology Planning \& Evaluation (IITP) grants funded by the Korea government (MSIP \& MSIT) (No. RS-2024-00443251, 40\%, Accurate and Safe Multimodal, Multilingual Personalized AI Tutors; No. RS-2024-00457882, 30\%, National AI Research Lab Project; No. IITP-2024-RS-2024-00436857, 30\%, Information Technology Research Center ITRC).
\bibliography{aaai25}

\clearpage

\appendix
\section*{Appendix}
In this supplement, we provide additional results demonstrating the practical construction of super-class hierarchies using similarity-based methods without predefined super-class hierarchies in Sec.~\ref{sup_sec:novel_super_class}. We also cover extending our knowledge transfer strategies to zero-shot HOI detection in Sec.~\ref{sup_sec:zero_shot_hoi}. Qualitative results are illustrated in Sec.\ref{sup_sec:qual_scr}, and further analyses and hyperparameter experiments are presented in Sec.\ref{sup_sec:experiments}. We include implementation details in Sec.~\ref{sup_sec:implementation details} and discuss the limitations in Sec.~\ref{sup_sec:limitation}.

\section{Predefined vs. Predicted Hierarchies}
\label{sup_sec:novel_super_class}
In this section, we aimed to demonstrate that hierarchical structures for super-classes can be effectively constructed using a simple similarity measure, even without prior knowledge of the super-classes.

\begin{figure}[ht]
    \centering
    \includegraphics[width=1.0\columnwidth]{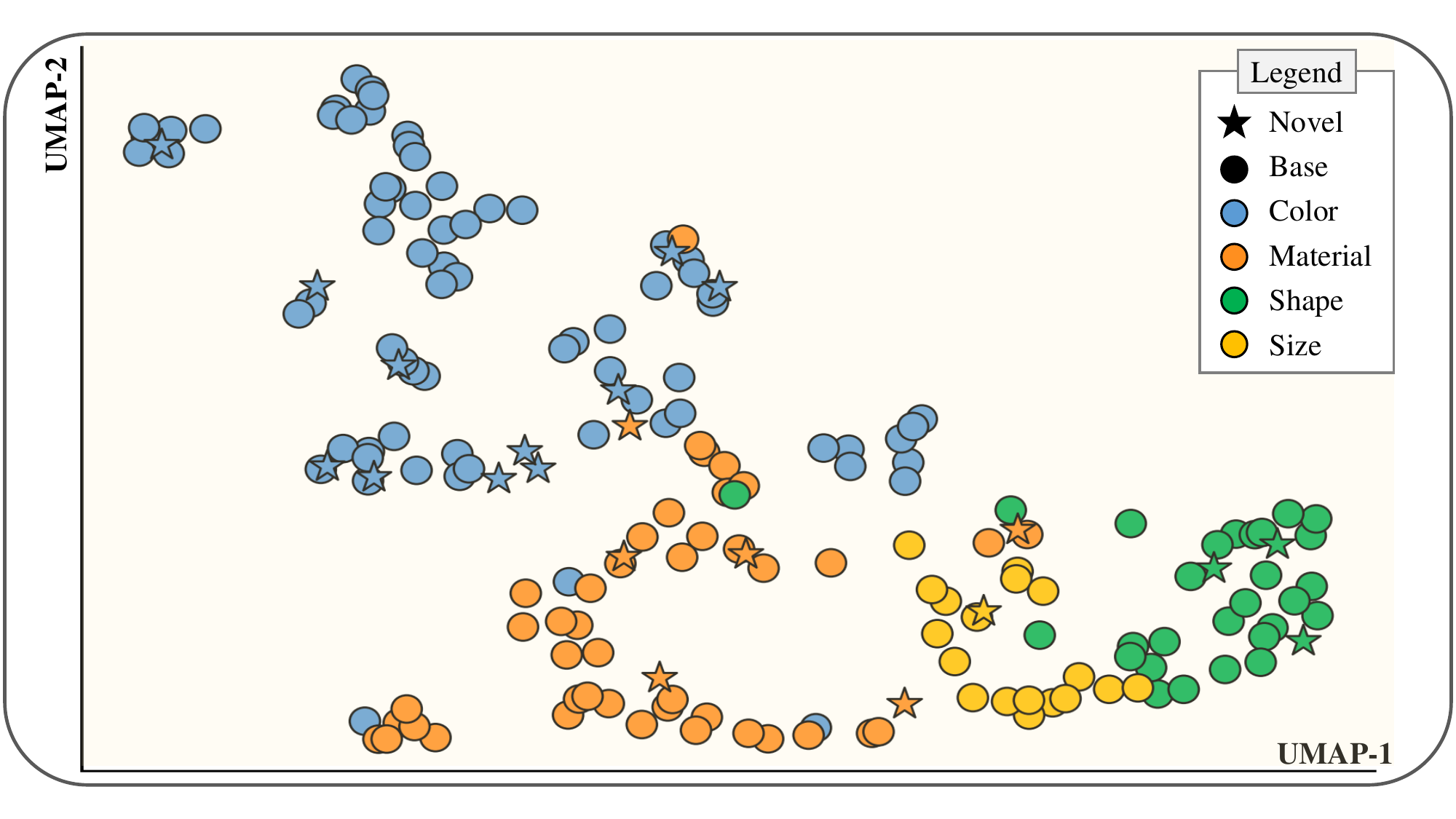}
    \caption{
    \textbf{Visusalization of attribute text embeddings.}
    The UMAP visualization shows class text embeddings. 
    The classes belonging to the same super-class are visualized with the same color and naturally form distinct, well-defined clusters.
    In addition, novel classes fall into the corresponding clusters.
    This explains why the similarity-based mapping function between classes and super-classes is effective.}
    \label{fig:UMAP_text_embedding}
\end{figure}

\begin{figure}[ht]
    \centering
    \begin{minipage}[t]{0.5\textwidth}
        \centering
        \includegraphics[width=0.8\textwidth]{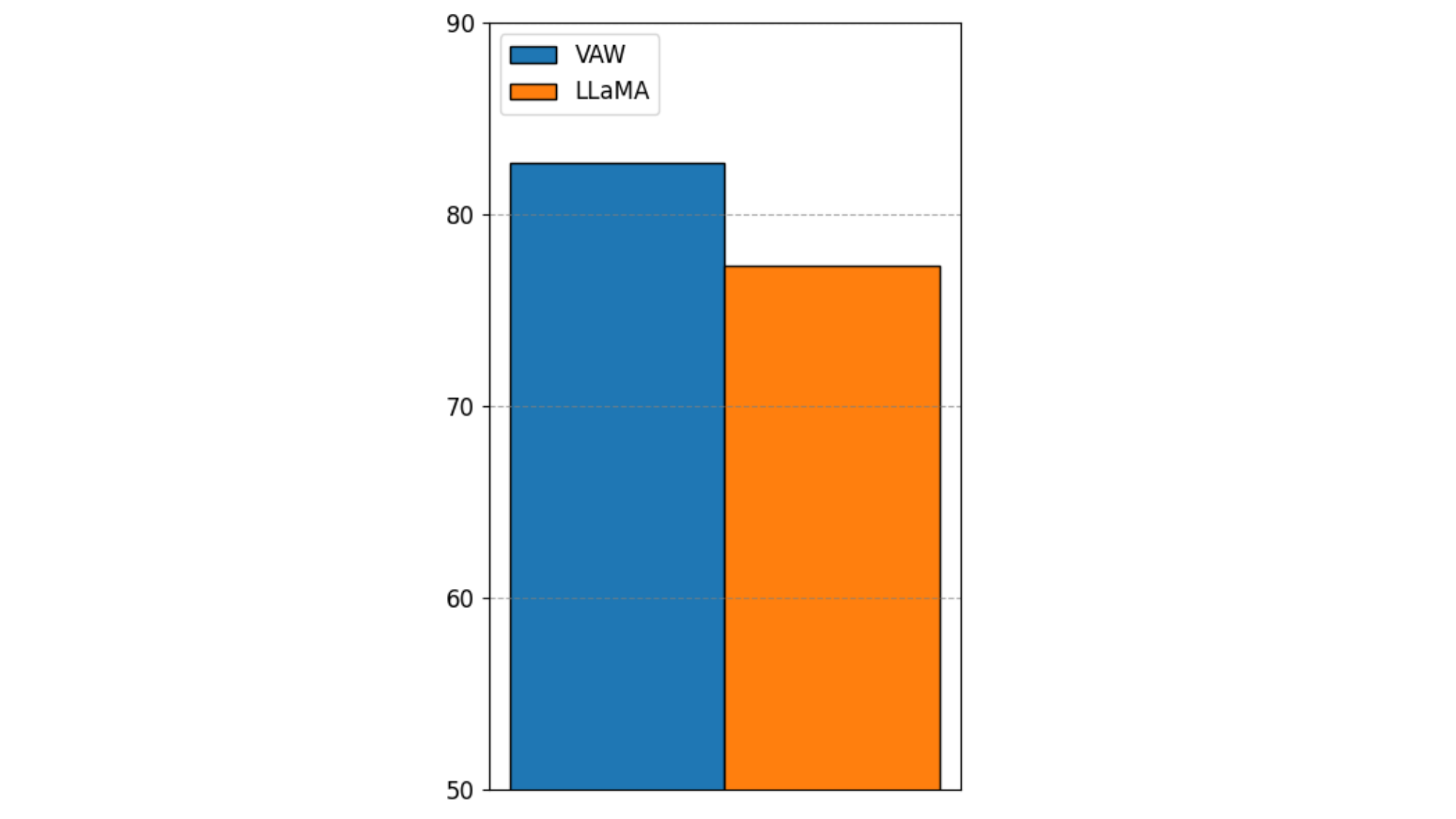}
        \vspace{-0.1cm}
        \subcaption{\textbf{Similarity-based super-class mapping.}
        }
        \label{fig:open-world-figure}
    \end{minipage}%
    \hfill
    \vspace{0.1cm}
    \begin{minipage}[t]{0.5\textwidth}
        \centering
        \begin{adjustbox}{width=0.8\textwidth}
        \begin{tabular}{c|ccc} 
          \toprule
          \textbf{Super-class mapping} &$\textbf{AP}_\textbf{base}$ & $\textbf{AP}_\textbf{novel}$ & $\textbf{AP}_\textbf{all}$ \\
          \midrule
          $\text{VAW}$ & 75.18 & 60.59 & 73.32 \\
          $\text{LLM}$ & 74.76 & 59.21 & 72.77 \\
          $\text{Similarity}$ & 74.95 & 59.34 & 72.96 \\
          \bottomrule
        \end{tabular}
        \end{adjustbox}
        \vspace{-0.0cm}
        \subcaption{\textbf{VAW zero-shot performance across hierarchical structures.}}
        \label{tab:open-world-table}
    \end{minipage}
    \vspace{-0.2cm}
    \caption{
    \textbf{Super-class mappings.} (a) demonstrates the effectiveness of our similarity-based super-class mapping by the cosine similarity with centroid embeddings to associate attributes with super-classes. 
    (b) shows the zero-shot performance of the model across different super-class mapping, including both predefined mapping in VAW and predicted super-class mappings by LLM or cosine similarity. The results support that our method can leverage the super-class queries without a ground truth mapping between attributes and super-classes.}
    \vspace{-0.6cm}
    \label{fig:combined}
\end{figure}

\noindent \textbf{Clustering and mapping of attribute embeddings.} 
Fig.~\ref{fig:UMAP_text_embedding} demonstrates that attribute text embeddings naturally form clusters by super-class, indicating that attributes within the same super-class have similar text embeddings.
Leveraging this, we propose a mapping function that uses centroid embeddings to build super-class hierarchies. We validated this by comparing our similarity-based method against two ground truths: (1) the VAW dataset's hierarchy and (2) a pseudo-hierarchy generated by a large language model (LLM)~\cite{touvron2023llama}. For each attribute text embedding, such as ``red'', we computed cosine similarities with centroid embeddings of super-classes, like ``color'', and classified the attribute into the super-class with the highest similarity.
As shown in Fig.~\ref{fig:open-world-figure}, our method achieved about 82.68\% accuracy for the VAW hierarchy and over 77.34\% for the LLM-generated hierarchy, with measurements excluding the dummy super-class ``other'', which lacks semantic meaning. \\

\noindent \textbf{Evaluating the effectiveness of hierarchy prediction.}
Additionally, Fig.~\ref{tab:open-world-table} shows that the VAW zero-shot performance with the LLM-generated hierarchy (second row) is almost identical to that with the predefined VAW hierarchy (first row). The third row represents the performance with a hierarchy constructed using the similarity-based mapping function. This consistency highlights the robustness of our similarity-based approach, demonstrating its effectiveness in constructing hierarchical structures for new attribute classes even without known super-classes.

\section{Extension to Zero-Shot HOI Detection}
\label{sup_sec:zero_shot_hoi}
\begin{table}[t]
    \begin{adjustbox}{width=\columnwidth}
    \begin{tabular}{c|c c|c|c c c}
        \toprule
        \textbf{Method} & \textbf{SCR} & \textbf{ZRSE} & \textbf{Type} & $\textbf{Seen}$ & $\textbf{Unseen}$ & $\textbf{Full}$ \\
        \midrule   
        \multirow{3}{*}{\text{UPT}} &  &  & \multirow{3}{*}{RF-UC} & 30.21 & 23.55 & 30.21 \\
        &\text{\checkmark} & & & 30.48 & 25.49  & 30.48  \\
        &\text{\checkmark} &\text{\checkmark} & & 32.28 &26.45 \textbf{\textcolor{red}{(+2.90)}} &31.12 \\
        \midrule   
        \multirow{3}{*}{\text{UPT}} &  &  & \multirow{3}{*}{\text{UV}} & 27.42 & 19.25 & 26.28 \\
        &\text{\checkmark} & &  & 27.86 & 21.74 & {27.01} \\
        &\text{\checkmark} &\text{\checkmark} &  & 28.59 & 22.19 \textbf{\textcolor{red}{(+2.94)}} & 27.70 \\
        \midrule
        \multirow{3}{*}{\text{PViC}} &  &  & \multirow{3}{*}{\text{RF-UC}} & 34.00 & 25.69 & 32.34 \\
        &\text{\checkmark} & &  & 34.15 & 27.11 & 32.74 \\
        &\text{\checkmark} &\text{\checkmark} &  & 34.45 & 28.16 \textbf{\textcolor{red}{(+2.47)}}& 33.19 \\
        \midrule   
        \multirow{3}{*}{\text{PViC}} &  &  & \multirow{3}{*}{\text{UV}} & 30.21 & 21.96 & 29.05 \\
        &\text{\checkmark} & &  & 30.05 & 23.19  & 29.09 \\
        &\text{\checkmark} &\text{\checkmark} &  & 30.39 & 23.94 \textbf{\textcolor{red}{(+1.98)}} & 29.48 \\
        \bottomrule
    \end{tabular}
    \end{adjustbox}
    \caption{\textbf{Extension to two-stage HOI detectors in Zero-Shot setting.}}
    \label{tab:hoi_extension}
\end{table}
In this section, we conduct an ablation study on the zero-shot HOI detection task to evaluate the effectiveness of knowledge transfer strategies.

\noindent \textbf{Dataset.} 
We conduct our experiments on the HICO-DET~\cite{chao2018learning} benchmark. HICO-DET contains 47,776 images, with 38,118 used for training and 9,658 for testing. The dataset includes annotations for 600 categories of HOI triplets, composed of 80 object categories and 117 action categories. Among these 600 HOI categories, 138 have fewer than 10 training instances and are defined as Rare, while the remaining 462 are described as Non-Rare.

\noindent \textbf{Evaluation metric.} 
We use mean Average Precision (mAP) as the evaluation metric. An HOI triplet prediction is considered a true positive if it meets the following conditions: 1) The Intersection over Union (IoU) between the predicted human and object bounding boxes and the ground truth bounding boxes exceeds 0.5; 2) The predicted interaction category is correct.

\noindent \textbf{Zero-shot construction.} 
We design our zero-shot experiments using two settings: Rare First Unseen Combination (RF-UC) and Unseen Verb (UV). The less frequent HOI categories are chosen as unseen classes in the RF-UC setting. The UV setting excludes certain action categories from the training set. For RF-UC, we select 120 HOI categories as unseen classes. In the UV setting, HOI categories involving 20 randomly chosen verb categories are excluded during training.

\noindent \textbf{Experiments.}
The baseline models used in the experiments were two-stage HOI detectors, incorporating UPT~\cite{zhang2022efficient} and PViC~\cite{zhang2023exploring}, which we reimplemented using BERT text embeddings with ResNet-50 backbone. For super-class guided consistency regularization (SCR), the \texttt{[MASK]} token feature obtained by prompting the ``action" super-class and the union box of human and object was used to apply pair-wise token regularization loss during training. Additionally, zero-shot retrieval-based score enhancement (ZRSE) was conducted in a training-free manner across 117 action classes with k=2.
As shown in Tab.~\ref{tab:hoi_extension}, for the UPT model in the RF-UC setting, the unseen mAP improved by 2.90 when both SCR and ZRSE were applied. In the UV setting, the unseen mAP increased by 2.94. Similarly, the PViC model showed an increase of 2.47 in the RF-UC setting and 1.98 in the UV setting when both methods were used.
These gains underscore the effectiveness of SCR and ZRSE in improving performance on unseen classes, demonstrating their value in enhancing generalization capabilities.

\section{Qualitative Results}
\label{sup_sec:qual_scr}
\begin{figure*}[htpb]
    \centering
    \includegraphics[width=1.0\textwidth]{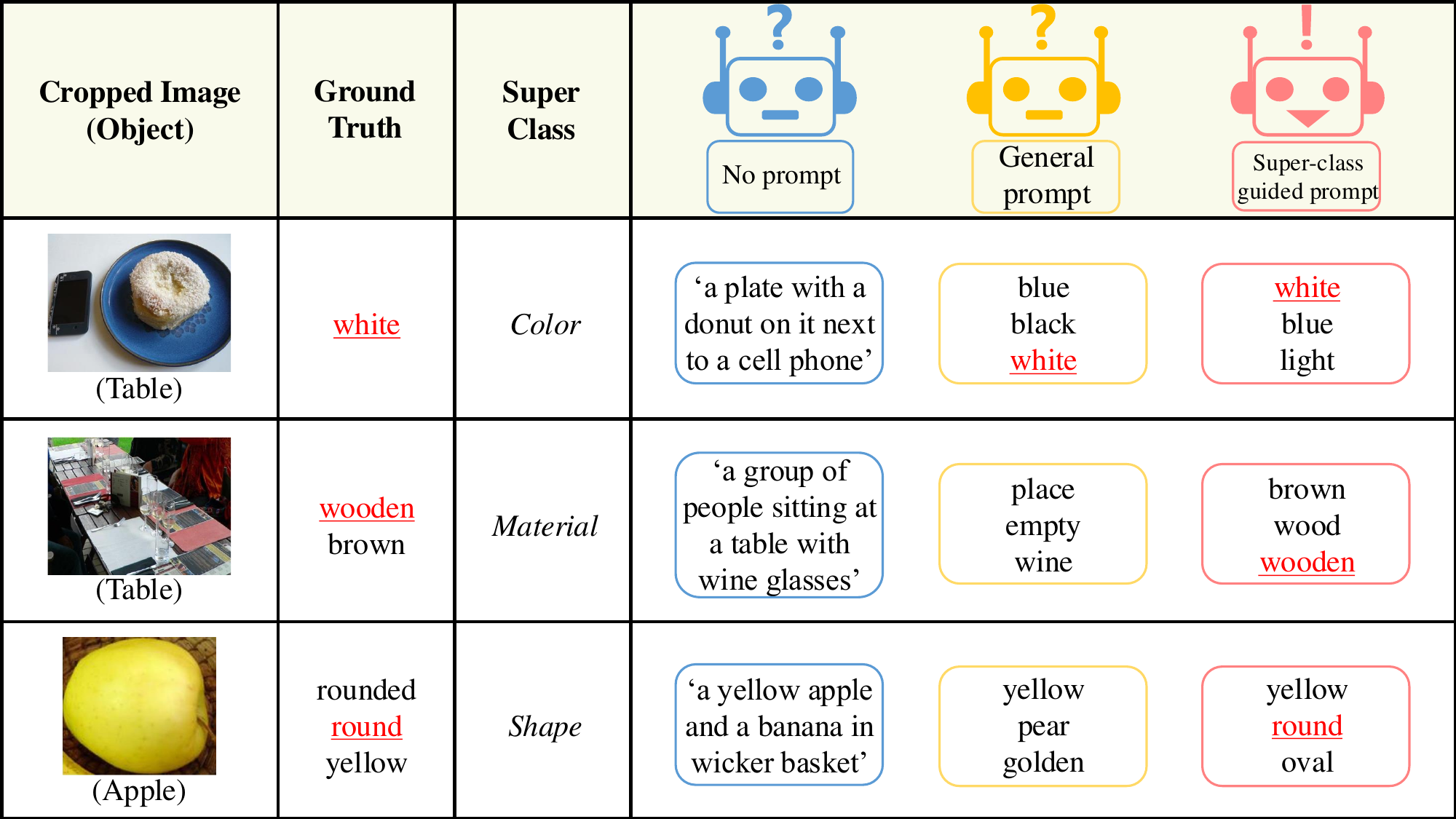}
    \centering
    \includegraphics[width=1.0\textwidth]{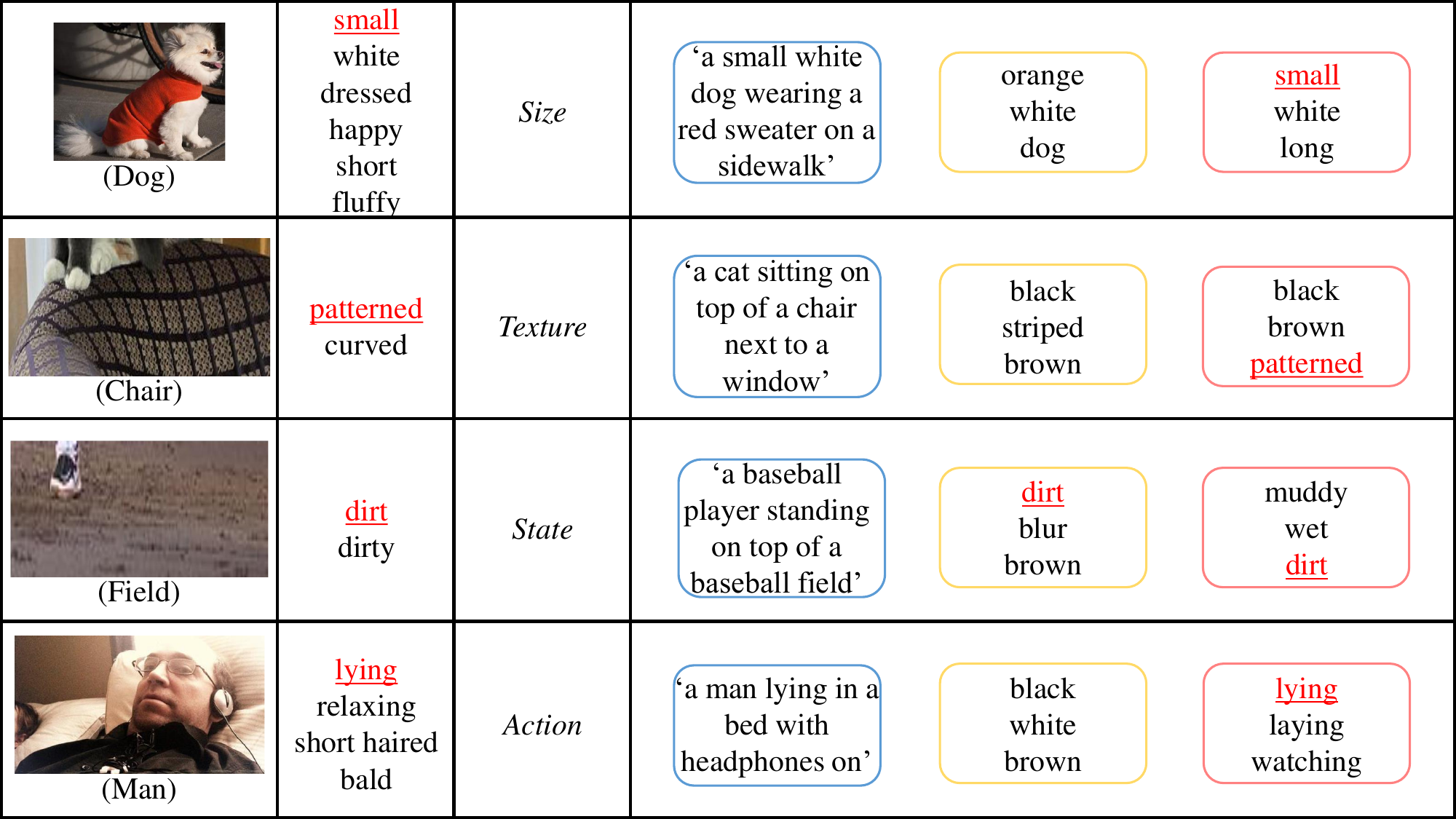}
    \vspace{0pt}
    \caption{\textbf{Qualitative results of generated caption and \texttt{[MASK]} token prediction from Q-Former.}
    }
    \label{fig:supple_figure}
\end{figure*}
In Section 3.3 of the main paper, we introduce a method called super-class guided consistency regularization (SCR). To evaluate the effectiveness of the \texttt{[MASK]} token's output, $\hat{\textbf{p}}^{l}$, we present some qualitative results. Fig.~\ref{fig:supple_figure} illustrates three types of outputs: 1) image captions generated by Q-Former, 2) attribute predictions from the \texttt{[MASK]} token using a super-class agnostic prompt: ``\texttt{The attribute of the [object category] is [MASK]}'', and 3) attribute predictions using a super-class specific prompt: ``\texttt{The [super class] of [object category] is [MASK]}''. The results show that when a super-class specific prompt is provided, the \texttt{[MASK]} token produces more accurate predictions. For example, in the second row, specifying the super-class leads to precise attribute predictions related to material. Similarly, in the third row, it is evident that specifying the super-class is crucial for accurate shape-related predictions.

\section{More experiments}
\label{sup_sec:experiments}
\renewcommand{\arraystretch}{1.1}
\begin{table}[t]
  \centering
  \begin{adjustbox}{width=\columnwidth}
  \begin{tabular}{ c |c |c c c}
    \toprule
    \multirow{1}{*}{\textbf{Method}}
     &\textbf{All} & \textbf{Head} & \textbf{Medium} &\textbf{Tail}\\
    \midrule
    LSEP~\cite{li2017improving}   &61.0 &69.1 &57.3 &40.9\\
    ML-GCN~\cite{chen2019multi}   &63.0 &70.8 &59.8 &42.7 \\
    Partial-BCE + GNN~\cite{pham2021learning}   &62.3 &70.1 &58.7 &40.1 \\
    SCoNE~\cite{pham2021learning}   &68.3  &76.5 &64.8 &48.0\\
    TAP~\cite{pham2022improving}  &73.4 &- &- &-\\
    GlideNet~\cite{metwaly2022glidenet} &71.2 &- &- &-\\
    \midrule    
    ML-Decoder$^\dagger$ ~\cite{ridnik2023ml} &72.1 &74.5 &71.7 &59.6 \\    
    SugaFormer &\textbf{74.2} &\textbf{77.0} &\textbf{74.5} &\textbf{62.5}\\
  \bottomrule
  \end{tabular}
  \end{adjustbox}
  \caption{\textbf{Results on the VAW in the fully-supervised setting.} $\dagger$ denotes the results obtained from our implementation.}
  \label{tab:main_supervised}
\end{table}

\begin{table}[t]
    \renewcommand{\arraystretch}{1.0}
    \centering
    \resizebox{0.33\textwidth}{!}{%
    \begin{tabular}{c |c c c}
        \toprule
        $\lambda$  &   
        $\textbf{AP}_\textbf{base}$ & $\textbf{AP}_\textbf{novel}$ & $\textbf{AP}_\textbf{all}$ \\
        \midrule                 
        0.1 & 74.72 & 57.41 & 72.51  \\
        1.0 & \textbf{75.26} & 60.13 & \textbf{73.33} \\
        2.0 & 75.18 & \textbf{60.59} & 73.32 \\
        3.0 & 74.48 & 59.11 & 72.52 \\
        \bottomrule
    \end{tabular}%
    }
    \caption{\textbf{Analysis on SCR coefficient $\lambda$}.}
    \label{tab:kd_coeff}
\end{table}

\begin{table}[t]
    \renewcommand{\arraystretch}{1.0}
    \centering
    \resizebox{0.33\textwidth}{!}{%
    \begin{tabular}{c |c c c}
        \toprule
        \textbf{TOP-K}  &   
        $\textbf{AP}_\textbf{base}$ & $\textbf{AP}_\textbf{novel}$ & $\textbf{AP}_\textbf{all}$ \\
        \midrule                 
        1  & \textbf{75.61} & 59.84 & \textbf{73.60} \\
        2 & 75.18 & \textbf{60.59} & 73.32 \\
        5 & 75.49 & 59.80 & 73.49 \\
        20 & 75.47 & 59.54 & 73.44 \\
        \bottomrule
    \end{tabular}%
    }
    \caption{\textbf{Analysis on ZRSE}.}
    \label{tab:zrse}
\end{table}

In this section, we present additional experimental results in the VAW~\cite{pham2021learning} in the both fully-supervised and zero-shot settings.

% \noindent \textbf{Fully-supervised setting on VAW.} Within a fully-supervised framework for attribute classification, we use mean average precision (mAP) to evaluate model accuracy, focusing on overall performance and class imbalance in the ``head'', ``medium'', and ``tail'' categories. As shown in Tab.~\ref{tab:main_supervised}, SugaFormer improves by 2.1 overall mAP  compared to the baseline, with increases of 2.5 mAP in the head classes, 2.8 mAP in the medium classes, and 2.9 mAP in the tail classes. These results highlight SugaFormer’s achievement of state-of-the-art performance in the fully-supervised setting and its significant improvement over the baseline in addressing class imbalance.

\noindent \textbf{Fully-supervised setting on VAW.} In the fully-supervised framework for attribute classification, we use mean average precision (mAP) to evaluate model performance, particularly focusing on overall accuracy and class imbalance across ``head'', ``medium'', and ``tail'' categories. As shown in Tab.~\ref{tab:main_supervised}, SugaFormer achieves a 2.1 mAP improvement overall compared to the baseline, with notable gains of 2.5 mAP in head classes, 2.8 mAP in medium classes, and 2.9 mAP in tail classes. These results demonstrate SugaFormer’s state-of-the-art performance and its effectiveness in addressing class imbalance, ensuring robust improvements across all attribute categories.

\noindent \textbf{Analysis on SCR coefficient $\lambda$.}
In our analysis on super-class guided consistency regularization (SCR), as shown in Tab.~\ref{tab:kd_coeff}, when $\lambda=1.0$, the highest performance is observed on $\text{AP}_\text{base}$ and $\text{AP}_\text{all}$. Meanwhile, when $\lambda=2.0$, the best performance is achieved in terms of $\text{AP}_\text{novel}$, highlighting its value in recognizing novel attribute classes.

\noindent \textbf{Analysis on ZRSE.}
We examined the influence of logit selection on results in our exploration of utilizing the \text{TOP-K} logits for zero-shot retrieval-based score enhancement (ZRSE), as detailed in Tab.~\ref{tab:zrse}. When we added the two highest logits during the inference, the score for $\text{AP}_\text{novel}$ was the best.

\noindent \textbf{Exploration on SQI.}
\begin{table}[t]
    \renewcommand{\arraystretch}{1.0}
    \centering
    \resizebox{0.33\textwidth}{!}{%
    \begin{tabular}{c c |c c c}
        \toprule
        \textbf{att} & \textbf{obj} &   
        $\textbf{AP}_\textbf{base}$ & $\textbf{AP}_\textbf{novel}$ & $\textbf{AP}_\textbf{all}$ \\
        \midrule                 
        &  & 74.96 & 58.53 & 72.87 \\
        &\checkmark  & 74.99 & 60.17 & 73.1 \\
        \checkmark &  & 74.54 & 59.40 & 72.61 \\
        \checkmark &\checkmark &\textbf{75.18} & \textbf{60.59} & \textbf{73.32} \\
        \bottomrule
    \end{tabular}%
    }
    \caption{\textbf{Effect of different pooling function in SQI}.}
    \label{tab:agg}
\end{table}
We explain our super-class query initialization (SQI) in detail and assess its effectiveness. 
Our initialization scheme combines the output embedding $\hat{\mathbf{z}}$ from Q-Former~\cite{li2023blip} with the text embedding of \text{j-th} super-class $\textbf{q}_{j}$.
For the initialization, we employ an pooling function $\textbf{Pool}(\cdot)$ to select the best output embedding among $\hat{\textbf{z}}$.
Given the text embedding of the object category $\textbf{t}_\textbf{o}$, detailed process of the selection is as follows:
\begin{equation}
    \begin{split}
    \hat{k}_\textbf{o} = \underset{k}{\mathrm{argmax}}{\;\textbf{t}_\textbf{o}\cdot \hat{\textbf{z}}_k^{\top}},
    \end{split}
    \label{eq:agg_detail}
\end{equation}
where $\hat{k}_\textbf{o}$ is an index of $\hat{\textbf{z}}$ which has the highest inner product value with $\textbf{t}_\textbf{o}$.
As shown in Tab.~\ref{tab:agg}, utilizing both the super-class text embedding and the output embedding to initialize the super-class query enhances the performance of the model.
Note that \textbf{att} and \textbf{obj} indicate the procedures for aggregating the output embeddings with the attribute class and object category embedding, respectively.
Specifically, the optimal outcomes were obtained when visual features derived from both object category embedding and a set of attribute class embeddings were employed.

\section{Further implementation details}
\label{sup_sec:implementation details}
\begin{figure*}[htpb]
    \centering
    \includegraphics[width=1.0\textwidth]{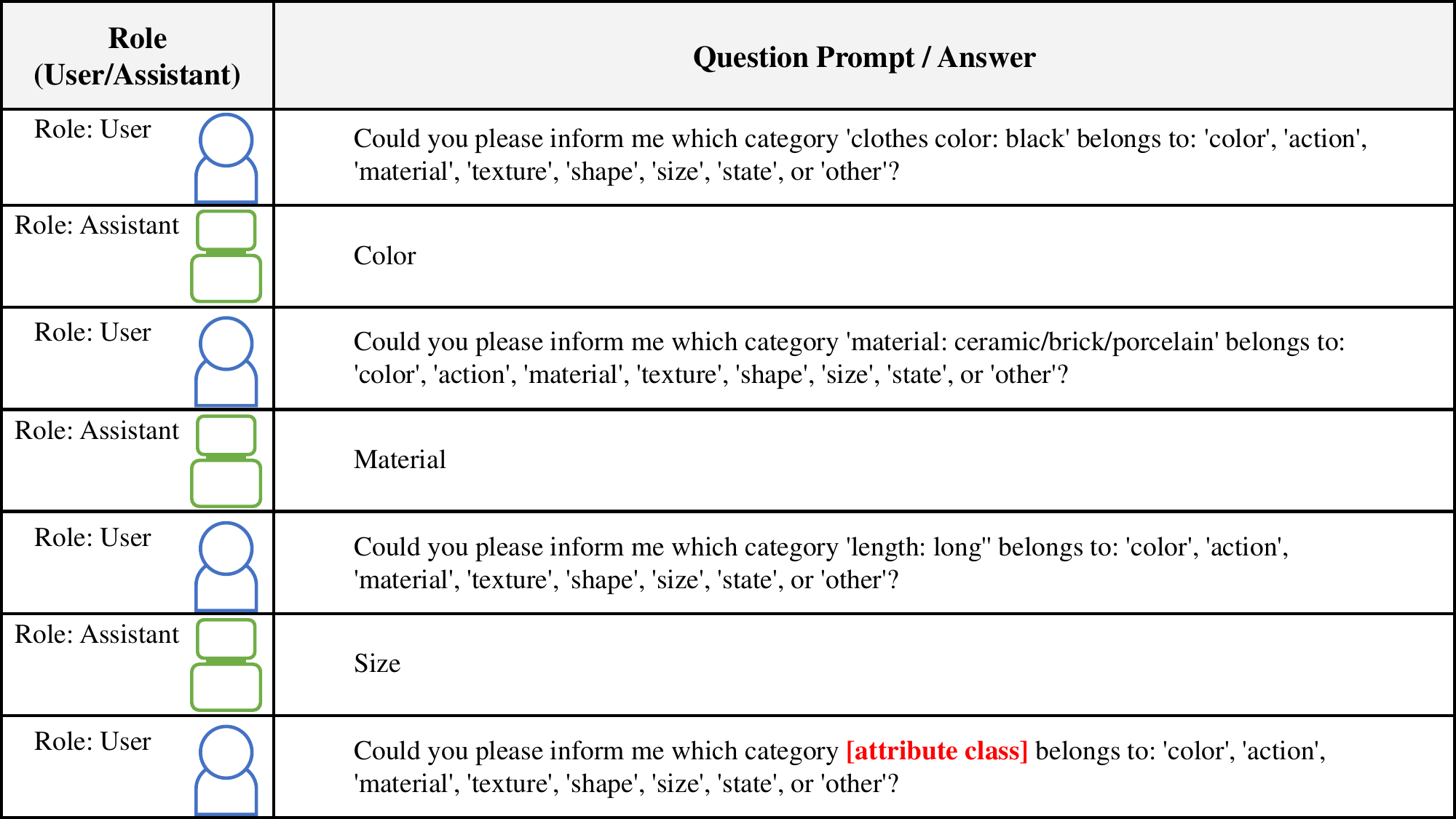}
    \caption{\textbf{Prompt for super-class pseudo-label.}
    }
    \label{fig:prompt_example}
\end{figure*}

In this section, we provide detailed explanations of the training process, model architecture, and the generation of super-class pseudo-labels for our experiments. \\
\noindent \textbf{Details of the training.}
% \subsection{Details of the Training}
In this research, we trained the VAW dataset using two settings: a fully-supervised setting for 15 epochs, starting with a learning rate of 1e-4 and reducing it to 1e-5 after the 13-th epoch, and a zero-shot setting for 9 epochs. Using a total batch size of 16. We used the AdamW optimizer with a weight decay of 1e-4. The model's prediction loss was determined using an asymmetric loss~\cite{ridnik2021asymmetric} with specific hyperparameters: $\gamma_{+}$ = 0, $\gamma_{-}$ = 4, and a clipping value of 0.05. 
When evaluating with OVAD, the dataset provides only the box coordinate and the class name of each object instance.
To generate segmentation masks and define super-classes, we utilize SAM~\cite{kirillov2023segment} and LLaMA~\cite{touvron2023llama} to generate a segmentation mask for each object and super-classes, respectively.

\noindent \textbf{Details of the model architecture.}
% \subsection{Details of the Model Architecture}
% TO DO dimension details
SugaFormer's decoder consists of three blocks, each comprising a cross-attention layer and a FFN. The overall hidden dimension of the decoder is set to 256, while the dimension within the FFN is 2048. For both attribute class embedding and super-class embedding, the \texttt{[CLS]} token embedding extracted using Q-Former was utilized, projected into a 256-dimensional space, and employed for training and testing purposes.

\noindent \textbf{Generating super-class pseudo-label.}
% \subsection{Generating super-class pseudo-label}
In the case of the OVAD~\cite{bravo2023open}, LSA~\cite{pham2022improving} dataset, not all datasets come with pre-defined super-classes. To address this, we exploit a large language model (LLM) to generate the super-class pseudo-label. Specifically, we employ LLaMA~\cite{touvron2023llama} along with a question prompt to determine the super-class to which the attribute class belongs. The question prompt used for this purpose is illustrated in Fig~\ref{fig:prompt_example}. Note that we utilize the pre-defined eight super-classes from the VAW dataset~\cite{pham2021learning}: color, action, material, texture, shape, size, state, or other. We induce the model to select one of these super-classes by leveraging a few-shot learning strategy.

\section{Limitation}
\label{sup_sec:limitation}
SugaFormer introduces a novel approach for zero-shot attribute classification using super-classes, but it has two limitations. First, We have effectively addressed the challenge of lacking predefined super-classes through text similarity-based mapping and LLM-based prediction, demonstrating that our method can leverage super-class queries without relying on ground truth mappings between attributes and super-classes. However, there may be even better mapping methods to explore that could further enhance performance.
Second, the model's structure can lead to a diminished learning signal when multiple classes within the same super-class are labeled, as they all contribute to the same query output, reducing the model's ability to distinguish between individual classes. Addressing these challenges offers promising avenues for future research.
\end{document}